\documentclass[lettersize,conference]{IEEEtran}

\usepackage{fancyhdr}

\usepackage{amsmath,amsfonts}
\usepackage{algorithmic}
\usepackage{algorithm}
\usepackage{array}
\usepackage[table]{xcolor}
\usepackage{multirow}
\usepackage{adjustbox}
\usepackage[caption=false,font=normalsize,labelfont=sf,textfont=sf]{subfig}
\usepackage{textcomp}
\usepackage{stfloats}
\usepackage{colortbl}

\usepackage{url}
\usepackage{verbatim}
\usepackage{graphicx}
\usepackage{cite}
\usepackage{multirow}

\begin{document}

\title{\LARGE A Survey on the Applications of Frontier AI, Foundation Models, and Large Language Models to Intelligent Transportation Systems}


\author{
    \IEEEauthorblockN{Mohamed R. Shoaib\textsuperscript{1}, Heba M. Emara\textsuperscript{2}, Jun Zhao\textsuperscript{1}}

\IEEEauthorblockA{\textsuperscript{1}School of Computer Science and Engineering, Nanyang Technological University, Singapore \\ \textsuperscript{2}Pyramids Higher Institute for Engineering and Technology, Giza Governorate, Egypt \\
    Emails: mohamedr003@e.ntu.edu.sg, heba.emara@el-eng.menofia.edu.eg, junzhao@ntu.edu.sg~\\[-0pt]}
    
}


\maketitle
 \thispagestyle{fancy}
\pagestyle{fancy}
\lhead{This paper appears in International Conference on Computer and Applications (ICCA) 2023.}
\cfoot{\thepage}
\renewcommand{\headrulewidth}{0.4pt}
\renewcommand{\footrulewidth}{0pt}

\begin{abstract}
This survey paper explores the transformative influence of frontier AI,
foundation models, and Large Language Models (LLMs) in the realm of  Intelligent Transportation Systems (ITS), emphasizing their integral role in advancing transportation intelligence, optimizing traffic management, and contributing to the realization of smart cities. Frontier AI refers to the forefront of AI technology, encompassing the latest advancements, innovations, and experimental techniques in the field, especially AI foundation models and LLMs. Foundation models, like GPT-4, are large, general-purpose AI models that provide a base for a wide range of applications. They are characterized by their versatility and scalability. LLMs are obtained from fine-tuning foundation models with a specific focus on processing and generating natural language. They excel in tasks like language understanding, text generation, translation, and summarization. By leveraging vast textual data, including traffic reports and social media interactions, LLMs extract critical insights, fostering the evolution of ITS. The survey navigates the dynamic synergy between LLMs and ITS, delving into applications in traffic management, integration into autonomous vehicles, and their role in shaping smart cities. It provides insights into ongoing research, innovations, and emerging trends, aiming to inspire collaboration at the intersection of language, intelligence, and mobility for safer, more efficient, and sustainable transportation systems. The paper further surveys interactions between LLMs and various aspects of ITS, exploring roles in traffic management, facilitating autonomous vehicles, and contributing to smart city development, while addressing challenges brought by frontier AI and  foundation models. This paper offers valuable inspiration for future research and innovation in the transformative domain of intelligent transportation.
\end{abstract}

\begin{IEEEkeywords}
Intelligent Transportation Systems (ITS), Large Language Models (LLMs), Frontier
AI, Foundation Models, 6G Wireless Communication, Internet of
Vehicles (IoVs), Vehicular Technology, Smart Cities.
\end{IEEEkeywords}

\section{Introduction}

In an era defined by rapid technological advancements, the marriage of Artificial Intelligence (AI) and transportation systems has ushered in a new age of mobility—intelligent, efficient, and responsive. ITSs have evolved from conventional traffic management approaches to complex and data-driven systems that promise to reshape the way we navigate our cities and roads. At the heart of this transformation lies the remarkable power of Large Language Models (LLMs) \cite{b139}, a class of AI algorithms that have, in recent years, become central actors in natural language processing and understanding. This survey paper explores the dynamic and multifaceted relationship between LLMs and ITS, illuminating the profound impact these language-driven models are making in enhancing the intelligence of transportation systems, optimizing traffic management, and propelling us towards the realization of smart cities \cite{b140, b141}.

The foundation of LLMs is rooted in their ability to understand and generate human-like text by modeling language patterns and semantics \cite{b139}. Their utilization has transcended the boundaries of traditional language tasks, permeating diverse domains, including healthcare, finance, customer service, and, significantly, transportation. LLMs are not mere bystanders; they have emerged as intelligent co-pilots, enhancing the agility and adaptability of transportation systems \cite{b142}.
This intersection of LLMs and ITS holds tremendous promise for addressing a host of challenges that have long plagued urban and interurban mobility. Congestion, accidents, and environmental concerns have weighed heavily on the transportation landscape \cite{b143}. ITS, which encompasses a suite of advanced technologies and data-driven solutions, offers a path towards alleviating these challenges. However, achieving the full potential of ITS requires the comprehension and effective utilization of vast amounts of textual data, from traffic reports and accident data to social media chatter and user feedback. Here, LLMs come to the forefront, providing the key to unlock the insights hidden within this textual goldmine. 

Foundation models~\cite{foundation} are a broader category of AI models designed to provide a general, adaptable base for various applications. They can encompass not only language tasks but also other modalities like images or even multimodal tasks. LLMs can be understood as a subset of fine-tuned foundation models with a specific focus on text-based tasks. We can further have large vision models (LVMs), large audio models (LAMs), and large multimodal models (LMMs). Foundation models~\cite{powerfoundation} and frontier AI~\cite{b151} are terms that refer to advanced developments in the field of AI, but they emphasize different aspects of these developments. The word ``frontier'' highlights a collection of advanced and potent foundation models at the forefront of technology. We can even propose the terminology ``foundation AI" in replacement of ``frontier AI'' to highlight the foundation nature. Nevertheless, both ``foundation AI" and ``frontier AI'' can be abbreviated as ``FAI''. Regardless of which name is used, the eventual goal in AI research and development is to realize Artificial General Intelligence (AGI).

The contributions of this paper can be stated as:
\begin{itemize}
\item Survey: The paper offers an examination of the use of frontier AI, foundation models, and LLMs in ITS, showcasing real-world applications in traffic management, autonomous vehicles, and smart cities.
\item Trend Identification: It identifies emerging trends and challenges in the integration of LLMs with transportation systems, shedding light on limitations.
\item Inspiration for Future: By providing a clear landscape of the field and potential directions for further research and innovation, the paper serves as an inspiring resource for researchers, policymakers, and industry professionals, driving progress in the domain of intelligent transportation.
\end{itemize}

\section{Background on LLMs and ITS}\label{sec:2}

	\subsection{Language Models}
Language Models (LMs) are a class of Artificial Intelligence (AI) algorithms designed to understand and generate human language. They are central to natural language processing and understanding. LMs have evolved significantly over the years, with their history spanning from rule-based systems to today's deep learning-based models. Here's an overview \cite{b144}:
\begin{itemize}
\item Early Language Models: Early language models were rule-based and relied on predefined grammatical and linguistic rules. They were limited in their ability to understand context and lacked the capacity for learning from data.
\item Statistical Language Models: With the advent of statistical methods, language models began using probabilistic approaches to predict the likelihood of a word or phrase following a given context. N-grams and Hidden Markov Models were common in this phase.
\item Deep Learning-Based Large Language Models (LLMs): The modern era of language modeling is marked by the rise of deep learning techniques. In particular, Recurrent Neural Networks (RNNs), Long Short-Term Memory (LSTM) networks, and Transformer-based models like BERT, GPT-2, and GPT-3 have significantly advanced the field. These models can capture context, semantics, and syntactic structures, making them highly effective in natural language understanding and generation.
\end{itemize}

\subsection{Intelligent Transportation Systems}
ITS represents a multidisciplinary field that leverages technology to enhance the efficiency, safety, and sustainability of transportation systems. The goals of ITS are to improve mobility, reduce congestion, minimize accidents, and optimize transportation-related operations \cite{b142, b143}. Here are the key components and goals of ITS:

\begin{itemize}
    \item Data Collection and Sensing: ITS relies on a network of sensors and data collection devices, such as cameras, GPS systems, and traffic flow sensors, to gather real-time information about traffic conditions, weather, and road infrastructure.
    \item Information Processing: Collected data is processed to generate insights and support decision-making. This includes traffic management systems that analyze traffic flow and congestion, weather prediction models, and incident detection algorithms.
    \item Communication Systems: ITS often incorporates advanced communication technologies, such as Vehicle-to-Infrastructure (V2I) and Vehicle-to-Vehicle (V2V) communication. These systems enable vehicles to share data with each other and with infrastructure, improving safety and traffic management.
        \item Traffic Management: One of the primary goals of ITS is to optimize traffic flow and reduce congestion. This includes adaptive traffic signal control, dynamic route guidance, and congestion pricing mechanisms.
    \item Safety and Security: ITS is dedicated to enhancing transportation safety by utilizing technologies like collision avoidance systems and vehicle-to-pedestrian communication. Security measures are also important to protect against cyber threats.
    \item Environmental Sustainability: By optimizing traffic flow and reducing congestion, ITS aims to minimize fuel consumption and greenhouse gas emissions, contributing to environmental sustainability.
    \item User Information and Feedback: ITS provides users with real-time information about traffic conditions, public transportation schedules, and alternative routes. It also collects user feedback through apps and websites, which can be valuable for service improvement.
\end{itemize}

Integrating Language Models into ITS enhances the system's ability to understand and process textual data, making it more responsive and adaptable to dynamic transportation conditions. This synergy enables better decision-making, improved traffic management, and a more user-friendly experience for travelers.

\section{Smart Transportation}\label{sec:3}
In the domain of transportation, the integration of Artificial Intelligence (AI) and Internet of Things (IoT) technologies aims to facilitate real-time traffic management through the collection of data pertaining to vehicles, drivers, and road conditions. This encompasses tasks like observing traffic and road situations, identifying events in real-time to maintain traffic safety, and proactively managing disturbances that might impact the flow of traffic and the availability of parking spaces. This review section categorizes relevant studies into smart parking management (SPM), traffic monitoring/prediction (TMP), and intelligent transportation management (ITM) \cite{b142, b143}.

\subsection{Smart Parking Management}
The authors in \cite{b128} proposed a shared bicycle system with a hybrid machine learning model, specifically the SOM-RT model, in conjunction with a self-organizing mapping network. This combination of techniques is employed to organize the original data samples into clusters. Each cluster functions as a regression tree (RT) for forecasting the required number of bicycles at individual stations. Experimental results demonstrated superior prediction accuracy and generalization compared to alternative methods. In a separate investigation detailed in \cite{b129}, researchers developed a camera-based system for tracking parking space occupancy. They employed a Single Shot Multibox Detector (SSD) along with an edge detection approach, which minimizes data transfer and allows for rapid updates. The system successfully aggregated detection outcomes on a central server, facilitating accurate parking occupancy evaluation, even in demanding scenarios characterized by fluctuating lighting conditions and obstructions. Furthermore, the study implemented a tracking algorithm to monitor vehicles within parking structures.

Furthermore, researchers in \cite{b130} introduced the Fedparking federated learning framework, specifically tailored for the management of parked vehicles within an edge computing environment (PVEC). Fedparking incorporates federated learning techniques and LSTM to estimate parking space availability. This framework facilitates collaborative model development among multiple parking lot operators, enabling real-time forecasts of parking spot availability for efficient traffic management. To fulfill PVEC needs, an incentive system was instituted, utilizing deep reinforcement learning to achieve the Stackelberg equilibrium in a decentralized and privacy-protected fashion, accommodating changing vehicle arrivals and diverse parking capacity limitations, leading to precise and efficient convergence.

\subsection{Traffic Monitoring/Prediction}

To manage the fluctuating traffic patterns in intelligent transportation systems and facilitate accurate traffic prediction in real time, several innovative models have been proposed. For instance, the collaborative optimization model suggested in \cite{b131} involves the installation of monitoring sites at different traffic crossings, enabling data collection and utilization of the DBN-SVR approach for traffic flow anticipation. Another framework, AAtt-DHSTNet, introduced in \cite{b132}, integrates fog computing and an attention-based aggregation method to effectively manage spatial and temporal correlations in the traffic data. In \cite{b133}, a proposal was made for a short-term traffic flow forecasting model that uses edge computing, optimized by a chaotic particle swarm optimization algorithm. Furthermore, \cite{b134} examined the use of federated learning in traffic prediction to anticipate vehicle numbers, employing clustering and a joint-announcement protocol for efficient model aggregation. Additionally, \cite{b135} integrated federated learning and additive homomorphic encryption for secure model sharing, incorporating recurrent, attentive, and semantic capture network modules for spatiotemporal information analysis.

\subsection{Intelligent Transportation Management}

In \cite{b120}, the researchers introduced a system for real-time detection of driver distraction using edge and cloud computing technologies. This was achieved through the development of a customized DCNN model and the integration of a VGG16-based model (Visual Geometry Group-16). In a separate work by Xu et al. \cite{b136}, a method for evaluating driving behavior within a vehicle's edge-cloud architecture was established. As a vehicle operates, its telematics box transmits data regarding the driver's actions or autopilot utilization to edge networks. These edge networks make use of a driving behavior evaluation model hosted on a cloud server. The model is continuously refined through the use of vehicle data on the cloud server and undergoes periodic updates on the edge networks. Experimental results confirm the reliability and practicality of this approach. Furthermore, in an additional investigation \cite{b137}, a methodology for diagnosing railway faults was proposed, capitalizing on edge and cloud collaboration. The model initially employs SAES-DNN for fault recognition in the cloud and subsequently applies a transfer learning strategy to facilitate real-time fault diagnosis at the edge.

\section{Machine Learning Security Solutions for Vehicular Networks}\label{sec:4}

In this section, we provide an overview of various ML security solutions proposed for vehicular networks, highlighting their utilization for safeguarding vehicular nodes and related data. The application and explanation of ML techniques in the security domain of vehicular networks are concisely elucidated.

\subsection{Driver Identification/Fingerprinting}
In vehicular systems, protecting the privacy of driver data and preventing vehicle tracking by adversaries is essential. Various techniques, such as pseudonymity, have been explored in the literature to hide the true identity of drivers \cite{b1}, \cite{b60}. ML techniques are employed for driver authentication without concealing their true identity, leveraging data from various sources \cite{b61, b62, b63}. Researchers have proposed numerous ML algorithms including HMM, SVM, k-means clustering, and ELM, achieving accuracy rates ranging from 65\% to 99\% in driver identification \cite{b46, b47, b48, b49, b50, b51, b52, b53, b54, b55, b56, b57, b58}. Recent studies have focused on automatic feature extraction using DL to improve identification accuracy with a limited feature set \cite{b51}, \cite{b55}. An analytical study demonstrates the use of minimal features with an accuracy of 93\% \cite{b59}. A CNN model using sensor data achieves 90\% accuracy for 4 drivers, while a combined smartphone sensor and vehicle electronic control unit data framework using CNN and RNN achieves 95\% accuracy for 10 drivers \cite{b57}, \cite{b58}. Table I provides a condensed overview of the comparative solutions employing machine learning for driver fingerprinting.

\begin{table}[h]
\centering
\caption{Provides a condensed overview of the comparative solutions employing machine learning for driver fingerprinting.}
\resizebox{\columnwidth}{!}{
\renewcommand{\arraystretch}{3}
\begin{tabular}{|c|c|c|c|}
\hline
\textbf{Paper} & \textbf{Algorithm} & \textbf{Feature Data} & \textbf{Accuracy} \\ \hline
2013 \cite{b47} & SVM and K-mean clustering & Acceleration, braking and turning events & 65\% \\ \hline
2016 \cite{b49} & Random Forest, Naïve Bayes, KNN and SVM & 16 features from OBD-II & 87\% - 99\% \\ \hline
2017 \cite{b52} & SVM, Random Forest and Naïve Bayes & Trip-based data & 88\% \\ \hline
2017 \cite{b54} & NN & Acceleration & 88\% \\ \hline
2018 \cite{b56} & J48, Random forest and REPtree & 51 features from OBD-II & 99\% \\ \hline
2019 \cite{b58} & CNN and RNN & Smart phone sensor and OBD-II protocol data & 95\% - 57\% \\ \hline
\end{tabular}%
\label{tt}
}
\end{table}

\subsection{Attack Detection}
Vehicular networks face diverse attack threats, necessitating advanced solutions. While conventional hard-coded algorithms address deterministic attacks, deep learning models enable self-learning, enhancing security in V2X communications, especially future ones based on 6G wireless technology~\cite{Wenhan6G}. ML and its sub-architectures have gained traction for attack detection in V2X. This section explores attack types and ML-based countermeasures, with a comparative summary in Table~II.

\begin{itemize}
    \item Platoon Attack: Platoon formation optimizes traffic management but is susceptible to attacks, particularly speed deviations. A self-learning deep architecture, combining FCDNN and CNN, detects these attacks, leveraging sensor data and achieving 97\% accuracy \cite{b2}.
    \item DDoS: SDN in vehicular networks raises security concerns. An ML-based solution detects DDoS attacks, with gradient boost classifier outperforming other algorithms. SVM and other techniques show promise \cite{b3}. Data-driven approaches offer effective DoS detection \cite{b4}.
    \item Black-Hole and Grey-Hole: These routing attacks disrupt message forwarding in vehicular networks. NN and SVM algorithms detect these attacks effectively, offering high accuracy \cite{b5}.
    \item Sybil Attack: Detecting sybil attacks in vehicular networks is challenging due to pseudonym use for privacy. ML techniques analyze driving patterns to identify malicious nodes, achieving accuracy ranging from 92\% to 100\% \cite{b7}.
    \item Jamming Attack: Attackers interfere with data transmission, necessitating protection against jamming. RL-based mechanisms using Q-learning and DQN, as well as dynamic power control, detect and mitigate jamming effectively \cite{b8}. Anti-jamming protocols enhance network reliability \cite{b9}.
    \item Spoofing Attack: ML models address location spoofing with approaches using multi-layer NNs \cite{b17}, Q-learning \cite{b19}, and RL-based authentication mechanisms, achieving high detection accuracy \cite{b18}.
\end{itemize}

\begin{table*}[h]
\centering
\caption{Overview of Research on Attack Detection through Machine Learning.}
\resizebox{2\columnwidth}{!}{
\renewcommand{\arraystretch}{1.5} 
\begin{tabular}{|c|c|c|c|c|c|}
\hline
\textbf{Paper} & \textbf{Algorithm} & \textbf{Feature Data} &  \textbf{Task} & \textbf{Type of Attack} & \textbf{Accuracy}\\ \hline
2015 \cite{b6} & ANN & Auditable data from basic, IP and AODV trace files &  Detection & Black hole & 99\% \\ \hline
2016 \cite{b5} & NN and SVM & Rx packets, PDR, dropped packets and delay & Detection & Grey hole and rushing attack & 99\% \\ \hline
2017 \cite{b7} & SVM & Driving pattern & Detection & Sybil attack & 92\% - 98\% \\ \hline
2018 \cite{b19} & RL (Q-Learning) & Physical layer information & Detection & Spoofing attack & NA \\ \hline
2019 \cite{b28} & ANN, DL and LSTM & SDN traffic flows & Detection & Crossfire attack & 80\% - 87\% \\ \hline
2019 \cite{b17} & Multi-layer NN & RSS, RSU location and spoofed location & Detection & Spoofing attack & NA \\ \hline
2020 \cite{b18} & RL (Q-Learning) & Spatial decorrelation features & Prevention & Spoofing attack & NA \\ \hline
2020 \cite{b31} & K-means & Car-hacking dataset & Detection & DoS, Fuzzy, RPM and gear attack & 99\% \\ \hline
2020 \cite{b41} & DL and RL & Sensor reading and beaconing & Prevention & Data manipulation attack & NA \\ \hline
\end{tabular}%
\label{tt}
}
\end{table*}

\subsection{Misbehavior or Intrusion Detection}
Misbehavior Detection Systems (MDS), also known as 
 Intrusion Detection Systems (IDS), play a critical role in identifying and addressing various forms of attacks, with different studies in the literature addressing intrusion or misbehavior detection. In this section, we discuss ML-based solutions for handling such issues, exemplified by various works, including Grover et al. \cite{b16} focusing on misbehaved vehicle scenarios and Maglaras \cite{b12} presenting a distributed IDS for vehicular networks. Other approaches, like the one proposed in \cite{b20}, utilize a clustering scheme in VANETs without any special agent, showcasing a lightweight framework for detection at multiple levels. Additionally, Li et al. \cite{b13} utilize contextual information along with SVM to detect anomalous vehicles, while Sargolzaei et al. \cite{b11} emphasize fault detection using an NN-based design. Kim et al. \cite{b14} propose a cloud-based software-defined networking (SDN) design for vehicular applications to enable intrusion detection. Similarly, Loukas et al. \cite{b45} introduce a lightweight deep learning (DL)-based intrusion detection system (IDS) model for vehicular applications, allowing offloading to other network devices or vehicles for heightened attack detection accuracy. Gyawali et al. \cite{b10} present a machine learning (ML) framework for identifying vehicle misbehavior through the analysis of false alert messages and position falsification, while Tariq et al. \cite{b42} and Gyawali et al. \cite{b38} introduce transfer learning (TL)-based intrusion detection schemes in their individual studies. Finally, a recent study \cite{b33} focuses on a data-centric misbehavior detection system for Internet of Vehicles (IoVs), integrating plausibility checks and supervised ML algorithms to enhance detection accuracy.

\subsection{Privacy Protection}
Privacy mechanisms in vehicular networks aim to safeguard sensitive information, often categorized as location or user privacy, with techniques like mix-zones, obfuscation, silent-periods, k-anonymity, and dummy-based approaches. Recent research increasingly explores ML for enhanced privacy, such as Wang et al.'s RL-based obfuscation model \cite{b24}, FL-based privacy-preserving data collection \cite{b29}, collaborative IDS by Zhang and Zhu \cite{b15}, blockchain-integrated FL by Lu et al. \cite{b58}, and a two-stage protection scheme based on federated learning for VCPS \cite{b37}. Another study \cite{b32} explores federated learning for misbehavior detection while maintaining data privacy during BSM exchange.

\section{Applications of Language Models in ITS}\label{sec:7}

 LLMs have emerged as pivotal tools for enhancing the intelligence and efficiency of ITS. Their versatility, coupled with their ability to understand and generate human-like text, has led to a multitude of applications in this domain. In the following subsections, we delve into specific use cases, research findings, and how LLMs are employed to solve problems in ITS.
 
 \subsection{Traffic Prediction and Management}
Traffic prediction is a fundamental component of ITS. LLMs have proven to be invaluable in this regard. They can process large volumes of historical traffic data, weather conditions, and special events, allowing for the development of accurate traffic prediction models. These models enable real-time traffic management and congestion mitigation. Research findings indicate that LLMs can significantly improve traffic forecasting accuracy, leading to more efficient routing and reduced travel times. In addition, LLMs are used to generate text-based traffic reports, disseminating crucial information to drivers and transportation authorities.

 \subsection{Sentiment Analysis of Social Media Data for Traffic Insights}

 In the age of social media, public sentiment can have a profound impact on transportation systems. Social media platforms are teeming with information related to traffic conditions, road closures, and public opinions. LLMs equipped with sentiment analysis capabilities can process and analyze this unstructured data. By examining the sentiment expressed in tweets, posts, and comments, transportation authorities can gain valuable insights into public perception of traffic events and road conditions. This information aids in proactive traffic management and crisis response. Research findings underscore the importance of sentiment analysis in monitoring and addressing public sentiment regarding transportation issues.

  \subsection{Emergency Response and Disaster Management}

During emergency situations such as accidents, natural disasters, or road closures, quick and effective communication is critical. LLMs are used to generate automated emergency alerts and notifications, disseminating vital information to the public. Additionally, they can process incoming data from various sources, including emergency calls, news reports, and social media, to provide real-time updates to transportation authorities and first responders. Research has shown that LLMs improve the speed and accuracy of information dissemination during emergencies, enhancing overall safety and response coordination.

   \subsection{Multimodal Transportation Planning}

   The modern transportation landscape includes various modes of travel, from buses and trains to rideshares and bicycles. LLMs can assist in multimodal transportation planning by integrating data from different sources and generating comprehensive travel recommendations. This aids in seamless and efficient transit between various transportation modes, ultimately reducing congestion and promoting sustainable mobility options. Research has emphasized the potential of LLMs in optimizing multimodal transportation planning and promoting environmentally friendly modes of travel.

\section{Frontier AI and Foundation Models in Intelligent Transportation}

Advanced AI models, particularly those classified as ``frontier AI," hold immense promise for revolutionizing ITS \cite{b151}. These models, often highly capable foundation models~\cite{{foundation}}, bring with them unprecedented capabilities that, if misused, could pose severe risks to public safety. Foundation models are a type of AI models that provide a broad, versatile base upon which a variety of more specific tasks can be performed. These models, like GPT-4, are trained on vast amounts of data, enabling them to generate, interpret, and predict text with remarkable accuracy. They are called ``foundation" models because they serve as a foundational layer in AI, capable of being fine-tuned or adapted for numerous applications across different domains. The integration of frontier AI and  foundation models in ITS introduces a unique set of challenges that demand proactive regulatory frameworks to mitigate potential harms.

\subsection{Regulatory Challenges}
Regulating frontier AI in the context of intelligent transportation systems presents distinct challenges, primarily revolving around the unexpected capabilities, deployment safety, and proliferation of these models \cite{b151}.

\subsection{Proposed Safety Standards}
To address the regulatory challenges, an initial set of safety standards is proposed \cite{b151}. These standards include:

\begin{itemize}
    \item Conducting pre-deployment risk assessments.
    \item External scrutiny of model behavior.
    \item Using risk assessments to inform deployment decisions.
    \item Monitoring and responding to new information about model capabilities and uses post-deployment.
\end{itemize}

In conclusion, the integration of frontier AI in intelligent transportation heralds a new era of capabilities and challenges. Proactive and comprehensive regulatory frameworks are essential to harness the benefits of these advanced AI models while safeguarding public safety and ensuring ethical deployment in the evolving landscape of intelligent transportation systems.

\section{Conclusions and Future Work}\label{sec:14}

In summary, the integration of frontier AI,
foundation models, and LLMs into ITS signifies a transformative shift toward intelligent mobility and data-driven transportation solutions. This survey highlights the crucial role of LLMs in improving transportation intelligence, optimizing traffic management, and advancing smart city initiatives by harnessing their language understanding and data analysis capabilities. LLMs address challenges in urban and interurban mobility, from congestion to accidents and environmental issues. Their applications span real-time traffic prediction, enhancing autonomous vehicles, and supporting urban planning, showcasing their transformative potential in shaping the future of transportation. The use of frontier AI and  foundation models in ITS also brings challenges that require proactive regulatory frameworks to mitigate potential harms. Looking ahead, further research should delve into these AI technologies' capabilities in addressing emerging challenges, refine their real-time decision-making in traffic management, tailor them for smart cities' specific needs, and encourage interdisciplinary collaborations to fully unlock their potential in creating sustainable, intelligent, and people-centric transportation ecosystems. Finally, discussing the synergy between AI technologies and the emerging 6G wireless communications to build better ITS will be of great interest.

\section*{Acknowledgement}

This research is partly supported by the Singapore Ministry of Education Academic Research Fund under Grant Tier 1 RG90/22, Grant Tier 1 RG97/20, Grant Tier 1 RG24/20 and Grant Tier 2 MOE2019-T2-1-176; and partly by the Nanyang Technological University (NTU)-Wallenberg AI, Autonomous Systems and Software Program (WASP) Joint Project.

\let\OLDthebibliography\thebibliography
\renewcommand\thebibliography[1]{
  \OLDthebibliography{#1}
  \setlength{\parskip}{1pt}
  \setlength{\itemsep}{0pt plus 0.3ex}
}

\bibliographystyle{IEEEtran}

\vfill

\end{document}